\pdfoutput=1

\documentclass[11pt]{article}

\usepackage[]{emnlp2021}

\usepackage{times}
\usepackage{latexsym}

\usepackage[T1]{fontenc}

\usepackage[utf8]{inputenc}

\usepackage{microtype}

\usepackage{amssymb}
\usepackage{tabularx}
\usepackage{amsmath}
\usepackage{stmaryrd}
\usepackage{enumitem}
\usepackage{multicol}
\usepackage{multirow}
\usepackage{booktabs}

\definecolor{darkgreen}{rgb}{0.0, 0.5, 0.0}

\usepackage{todonotes}  

\title{An Analysis of Euclidean vs. Graph-Based Framing for \\ Bilingual Lexicon Induction from Word Embedding Spaces}

\author{Kelly Marchisio\textsuperscript{1}, Youngser Park\textsuperscript{4,5}, Ali Saad-Eldin\textsuperscript{3}, Anton Alyakin\textsuperscript{2}, \\ 
  \bf{Kevin Duh\textsuperscript{1,5}, Carey Priebe\textsuperscript{2,5}, Philipp Koehn\textsuperscript{1}} \\
        Depts. of \textsuperscript{1}Computer Science, \textsuperscript{2}Applied Mathematics and Statistics, and
        \textsuperscript{3}Biomedical Engineering \\ 
        \textsuperscript{4}Center for Imaging Science,
        \textsuperscript{5}Human Language Technology Center of Excellence\\ 
        Johns Hopkins University \\ 
        {\tt \{kmarc,youngser,asaadel1,aalyaki1\}@jhu.edu} \\
        {\tt kevinduh@cs.jhu.edu, \{cep, phi\}@jhu.edu}}

\begin{document}
\maketitle
\begin{abstract}
Much recent work in bilingual lexicon induction (BLI) views word embeddings as vectors in Euclidean space.  As such, BLI is typically solved by finding a linear transformation that maps embeddings to a common space. Alternatively, word embeddings may be understood as nodes in a weighted graph.  This framing allows us to examine a node's graph neighborhood without assuming a linear transform, and exploits new techniques from the graph matching optimization literature. These contrasting approaches have not been compared in BLI so far. In this work, we study the behavior of Euclidean versus graph-based approaches to BLI under differing data conditions and show that they complement each other when combined. We release our code at \url{https://github.com/kellymarchisio/euc-v-graph-bli}.

\end{abstract}

\section{Introduction}
Bilingual lexicons are useful in many natural language processing tasks including constrained decoding in machine translation, cross-lingual information retrieval, and unsupervised machine translation. There is a large literature inducing bilingual lexicons from cross-lingual spaces. ``Mapping" methods based on solving the orthogonal Procrustes problem and its generalizations are popular, where languages are mapped to a common space from which a lexicon is extracted. This has been successful when word embedding spaces are roughly isomorphic, but fails as embedding spaces diverge \cite{sogaard-etal-2018-limitations, vulic-etal-2019-really}. 

Rather than word embeddings in Euclidean space, we can work with weighted graphs derived from embeddings. 
Graphs may be full-connected or sparse to capture the underlying data manifold. For instance, we may create a similarity graph with words as nodes and cosine distance between word vectors as edges. 
The use of graphs in NLP has a rich history, for tasks as varied as summarization, part-of-speech tagging, syntactic parsing, information extraction, measures of semantic similarity, and evaluation of cross-lingual word embeddings \cite{mihalcea2011graph, nastase2015survey, fujinuma-etal-2019-resource}. Graphs can also represent rich relationships like hyponym/hypernym, syntactic roles, synonymy such as in WordNet \cite{miller1995wordnet} and Freebase \cite{freebase}. We focus on fully-connected graphs derived from pairwise cosine similarities between word embeddings.

\begin{figure*}[t]
  \centering
  \includegraphics[width=1\linewidth]{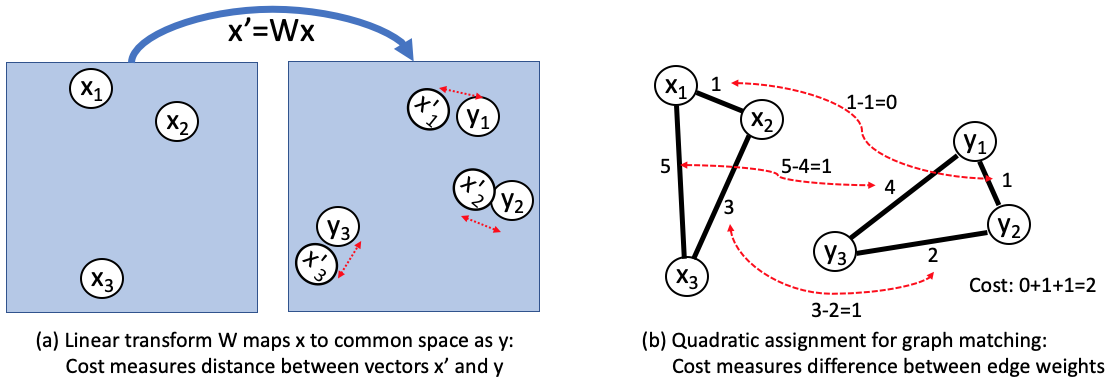}
  \caption{Comparing \textbf{Euclidean (Procrustes)} vs. \textbf{Graph (SGM)} views. (a) Euclidean view assumes common embedding space and computes costs based on pairs of vectors. (b) Graph-based view assumes graph structure and computes cost based on pairs of edges. Both exploit within-language neighborhood info but in different ways.}
  \label{fig:euc-vs-graph}
\end{figure*}

The Euclidean view, exemplified by methods solving the Procrustes problem, works with embedding spaces and assumes the existence of a linear transform that maps the spaces. The graph-based view works with graphs for each language and directly performs matching on edge pairs based on neighborhood information. This view is exemplified by graph matching methods that solve the quadratic assignment problem from the combinatorial optimization literature. \citet{ruder-etal-2018-discriminative} and \citet{haghighi-etal-2008-learning} incorporate related techniques for bilingual lexicon induction. We use Seeded Graph Matching \cite[SGM;][]{fishkind2019seeded} as representative of this class of approach. Figure \ref{fig:euc-vs-graph} illustrates the differences between the framings; while they both exploit the idea that words with similar neighbors (in Euclidean or graph space) should be translations of one another, they implement the idea in very different ways. 

We explore these two different views of BLI. Our main contributions are (a) a thorough comparison of Euclidean vs. graph-based framings to BLI under varying data conditions, and (b) a method for combining both approaches that achieves better performance than either alone.  

We organize our work into three main experimental setup and results sections. First, we compare standard algorithms of performing BLI via solutions to the orthogonal Procrustes problem (``Procrustes", for short) and SGM in Section \ref{noniter-exps}; we find that their performance varies depending on the number of seeds. SGM appears better when using less seeds.
Second, as it is common to improve results by bootstrapping, we compare iterative versions of Procrustes and SGM in Section \ref{iter-exps}. We find that Iterative Procrustes improves much more rapidly than Iterative SGM. We also introduce stochastic variants of the iterative algorithms to improve robustness and experiment with active learning setups. Finally, we present our combined system which outperforms individual Procrustes and SGM approaches in Section \ref{combo-exps}.

\section{Background}
\label{background}

BLI begins with two word embedding matrices: 
$\bf{X} \in \mathbb{R}^{n \times d}$ represents the $d$-dimensional word embeddings for $n$ vocabulary items in language X, 
and $\bf{Y} \in \mathbb{R}^{m \times d}$ represents the $m$ embeddings separately trained on monolingual data in language Y. 
We assume seeds $\{(x_1, y_1), (x_2, y_2),... (x_s, y_s)\}$ are given, which are supervised labels indicating translation correspondence between vocabulary items in the languages. 
We sort the corresponding submatrices of $\bf{X}$ and $\bf{Y}$ so each row of   $\overline{\bf{X}} \in \mathbb{R}^{s \times d}$ and $\overline{\bf{Y}} \in \mathbb{R}^{s \times d}$ corresponds to the seeds.  
Usually, $s$ is strictly smaller than both $n$ and $m$ and the goal is to find translation correspondences in the remaining words. 

\paragraph{Procrustes and linear transforms:} The popular Procrustes-based methods for BLI \cite[e.g.][]{artetxe2016learning, artetxe-etal-2019-effective, conneau-lample-2018, patra-etal-2019-bilingual} match seeds by calculating a linear transformation $\bf{W}$ by a variant of the below: 
\begin{equation}
\label{proc-eq}
    \min_{\bf{W}\in\mathbb{R}^{d \times d}} ||\overline{\bf{X}}\bf{W}-\overline{\bf{Y}}||_F^2
\end{equation}
If $\bf{W}$ is required to be orthogonal, then distances between points are unchanged by the transform and a closed form solution can be computed by singular value decomposition \cite{schonemann1966generalized}. 

Once languages are mapped to the same space by $\bf{W}$, nearest neighbor search finds additional translation pairs. 
If $\bf{W}$ is known, one can find translations by optimizing over permutations $\Pi$:
\begin{equation}
\label{eq:procrustes-permute}
    \min_{\bf{P}\in \Pi} ||\bf{X}\bf{W}-P\bf{Y}||_F^2
\end{equation}
$\bf{P} \in \{0,1\}^{n \times n}$ is permutation matrix that shuffles the rows of $\bf{Y}$. 
If we enforce the 1-to-1 correspondence, 
this is linear assignment problem that is solvable in polynomial time, e.g. with the Hungarian algorithm \cite{kuhn1955hungarian} or Wasserstein methods \cite{grave2019unsupervised}. 
In the NLP literature, a large number of methods are based on the same underlying idea of linear transform followed by correspondence search/matching (see Related Work). 

To extract lexicons, one performs nearest neighbor search on the transformed embeddings. To mitigate the hubness problem (where some words are close to too many others) \cite{radovanovic2010hubs, suzuki-etal-2013-centering}, \citet{conneau-lample-2018} modifies the similarity using cross-domain similarity local scaling (CSLS) to penalize hubs. 
For $x, y$ in embedding space $V$: 
\[
\text{CSLS}(x,y) = 2\cos(x,y) - \text{avg}(x, k) - \text{avg}(y, k)
\]
\[
\text{avg}(v, k) = \frac{1}{k}\sum\limits_{v_n\in N_k(v, V)}cos(v_n, v)
\]
$N_k(v, V)$ returns the k-nearest-neighbors to $v \in V$ by cosine similarity (typically $k=10$). 

\paragraph{Graph matching:} In fields such as pattern recognition, network science, and computer vision, there exist a large body of related work termed ``graph matching."
Rather than assuming the existence of a linear transform between the embedding spaces, these methods start with or construct two graphs and try to match vertices such that neighborhood structure is preserved. 
Intuitively, the motivation of preserving neighborhood structure is the same as Procrustes methods, but the absence of linear transform $\bf{W}$ is an important distinction that potentially makes graph matching more flexible. Indeed, some recent BLI work argue against linear transforms \cite{mohiuddin-etal-2020-lnmap} and discuss the failure modes due to lack of isometry \cite{sogaard-etal-2018-limitations,nakashole-flauger-2018-characterizing,ormazabal-etal-2019-analyzing, glavas-etal-2019-properly, vulic-etal-2019-really, patra-etal-2019-bilingual, marchisio-etal-2020-unsupervised}. 

For BLI, we may build the graphs as $\bf{G_x} = \bf{XX^T}$ and $\bf{G_y} = \bf{YY^T}$. 
For standard graph matching objectives, we restrict the vocabularies of $\bf{X}$ and $\bf{Y}$ to equal size, thus $\bf{G_x}, \bf{G_y} \in \mathbb{R}^{n \times n}$. 
We find the optimal relabeling of nodes such that:
\begin{equation}
    \min_{\bf{P}\in \Pi} ||\bf{G_x} -\bf{P G_y P}^T||_F^2
\end{equation}
This is an instance of the quadratic assignment problem and is much harder than Eq.~\ref{eq:procrustes-permute}. 
It is NP-Hard \cite{sahni1976p} but various approximation methods exist.
\citet{vogelstein2015fast} use the Frank-Wolfe method \cite{frank1956algorithm} to find an approximate doubly-stochastic solution, then project onto the space of permutation matrices.

When seeds are available, SGM can be applied to solve the amended objective in Equation \ref{sgm-opt}, where $s$ is the number of seeds and $\Pi_{n-s}$ is the set of permutation matrices for the $n-s$ non-seed words. See Appendix for details.
\begin{equation}
\label{sgm-opt}
\underset{\bf{P} \in \bf{\Pi_{n-s}}}{\min}\lVert \bf{G_x} - (\bf{I_s} \oplus \bf{P})\bf{G_y}(\bf{I_s} \oplus \bf{P})^T \rVert^2_F
\end{equation}

\subsection{Differences between Procrustes and SGM}
Two differences in behavior of Procrustes vs. SGM are worth discussing for their relevance to BLI.

\paragraph{\textit{Procrustes is many-to-one; SGM is one-to-one.}}
After solving the orthogonal Procrustes problem, translation pairs are selected by finding the $y \in \bf{Y}$ that is closest to the mapped source word in $x_w \in \bf{XW}$. 
It is possible that the nearest neighbor to both $x_{w_1} \in \bf{XW}$ and $x_{w_2} \in \bf{XW}$ may be $y_1 \in \bf{Y}$, so $\{(x_{w_1}, y_1), (x_{w_2}, y_1)\}$ may be induced as final translation hypotheses. 
Conversely, SGM solutions are strictly one-to-one; If $x_{w_1}$ is paired with $y_1$, then $x_{w_2}$ cannot be. 
As such, SGM may avoid hubs naturally without CSLS. 
A way around the one-to-one restriction is to use SoftSGM. 
For instance, if $x_{w_1}$ is paired with $y_1$ on 40\% of internal runs of SoftSGM and $x_{w_2}$ is paired with $y_1$ on 40\% of runs (and $y_1$ is the most frequent pairing for both $x_{w_1}$ and $x_{w_2}$), we may induce $\{(x_{w_1}, y_1), (x_{w_2}, y_1)\}$ as final hypotheses.

\paragraph{\textit{Procrustes is soft-seeded; SGM is hard-seeded.}} 
Procrustes is ``soft-seeded"; giving seed $(x_1, y_1)$ does not guarantee that $x_1$ and $y_1$ will be paired in the solution, because $y_1$ may not be the nearest neighbor to the mapped $x_{w_1}$. Conversely, SGM is ``hard-seeded": pairings given as seeds will always appear in the solution. This is ideal when one is confident about the quality of the seeds, but means that SGM is not robust to errors in the seed set.

\section{Experimental Setup}
Because there are three methods and results sections, we detail the experimental setup first. 
We evaluate on English$\rightarrow$German (En-De) and Russian$\rightarrow$English (Ru-En).
\paragraph{Monolingual Word Embeddings}
We use 300-dimensional monolingual word embeddings trained on Wikipedia using fastText \cite{bojanowski2017enriching}.\footnote{https://fasttext.cc/docs/en/pretrained-vectors.html}
We normalize to unit length, mean-center, and renormalize, following \citet{artetxe2018robust} (``iterative normalization", \citet{zhang-etal-2019-girls}).
\paragraph{Data \& Software}
Bilingual dictionaries from MUSE\footnote{https://github.com/facebookresearch/MUSE} are many-to-many lexicons of the 5000 most-frequent words from the source language, paired with one or more target-side translations. We filter each lexicon to be one-to-one for simplicity of analysis. For source words with multiple target words, we keep the first occurrence. This is equivalent to randomly sampling a target sense for polysemous source words because target words are in arbitrary order. En-De originally contains 14667 pairs, and 4903 remain after filtering. Ru-En has 7452 pairs, reduced to 4084. We use 100-4000 pairs as seeds, chosen in frequency order. The rest are the test set. Seed/test splits are in Table \ref{test-size}. We use the public implementation of SGM with random initialization from Graspologic\footnote{https://github.com/microsoft/graspologic} \cite{chung2019graspy}. We leave all other hyperparameters as their defaults (maximum Franke-Wolfe iterations: 30 with epsilon stopping criterion = 0.03, shuffle\_input=True).

\begin{table}[ht]
  \centering
  \begin{tabular}{@{}l|r@{\:\:\:}r@{\:\:\:}r@{\:\:\:}r@{\:\:\:}r@{\:\:}r@{}}
    \toprule
     Seeds & 100 & 200 & 500 & 1000 & 2000 & 4000 \\
     \midrule
    En-De Test & 4803 & 4703 & 4403 & 3903 & 2903 & 903 \\
    Ru-En Test & 3984 & 3884 & 3584 & 3084 & 2084 & 84 \\
    \bottomrule
  \end{tabular}
  \caption{Seed/test set splits for En-De, Ru-En.}
  \label{test-size}
\end{table}
\section{Non-Iterative Experiments}
\label{noniter-exps}
\subsection{Methods}
\paragraph{Procrustes} We compare Procrustes versus SGM methods when each is run once. We solve the orthogonal Procrustes problem of Equation \ref{proc-eq} over known seeds and apply the linear transform $\bf{W}$ to the entire source embedding matrix $\bf{X}$. For each mapped source word in $\bf{XW}$, we select $y$ from target embedding matrix $\bf{Y}$ with the minimum CSLS score as the translation.
\paragraph{SGM} We construct graphs $\bf{G_x} = \bf{XX^T}$ and $\bf{G_y} = \bf{YY^T}$, which are matrices of cosine similarity. We solve Equation \ref{sgm-opt} using the SGM algorithm from \citet{fishkind2019seeded}. We implement \citet{fishkind2019seeded}'s SoftSGM algorithm by running SGM ten times, each time using a different random initialization for the permutation matrix $\bf{P}$. This gives a probability distribution over matches. 
\paragraph{}
Standard metrics for BLI are precision@1 and precision@5 (p@1, p@5). Evaluating p@1 is straightforward. For Procrustes p@5, we select the five nearest neighbors per source word. Because SGM only makes one guess per source word, we calculate p@5 using SoftSGM \cite{fishkind2019seeded}, which returns a probability distribution over possible matches given multiple runs of SGM. We select the top five hypotheses per source word from the probability distribution.\footnote{There may be less than five hypotheses available per source word if there is not great diversity in output hypotheses.} We calculate recall@5 and F1@5 analogously.  

\subsection{Results}
Table \ref{tab:single-results} shows non-iterative results. SGM outperforms Procrustes in nearly all scenarios, and the effect with less seeds is particularly marked: Procrustes scores just 4.1\% with 100 seeds and 16.6\% with 500 seeds for Ru-En, while SGM scores 50.1\% and 52.2\%, respectively. With a moderate number of seeds, Procrustes and SGM perform similarly.\footnote{SoftSGM performs similarly to SGM, so is not reported.}
\begin{table}[ht]
  \centering
  \begin{tabular}{@{}r|cc|cc@{}}
    \toprule
    & \multicolumn{2}{c}{\underline{En-De}} & \multicolumn{2}{c}{\underline{Ru-En}} \\
    Seeds & Procrustes & SGM & Procrustes & SGM \\
    \midrule
    100 &  3.6 & \textbf{45.8} &  4.1 & \textbf{50.1} \\
    200 &  16.1 & \textbf{47.3} &  16.6 & \textbf{52.2} \\
    500 &  44.9 & \textbf{51.9} &  45.3 & \textbf{56.0} \\
    1000 & \textbf{57.2} & 54.9 & 56.6 & \textbf{58.1} \\
    2000 & \textbf{63.1} & 61.5 & 62.7 & \textbf{67.1} \\
    4000 & 70.8 & \textbf{74.2} & 67.9 & \textbf{89.3} \\
    \bottomrule
  \end{tabular}
  \caption{P@1 of Procrustes vs. SGM. 
  }
  \label{tab:single-results}
\end{table}

We evaluate p@5, recall@5, and F1@5 in Table \ref{tab:p5-single-results}. SGM has considerably higher precision and F1 than Procrustes across all experiments (by 50+ percentage points in extreme cases) but Procrustes generally has greater recall when seed size is 500 or greater. With 100 or 200 seeds, SGM outperforms Procrustes across-the-board. We note the difference in the number of translation hypotheses induced for each method in ``Total Hyps."
\begin{table*}[ht]
  \centering
  \begin{tabular}{@{}r@{ }r|r@{  }r|r@{  }r|r@{  }r|r@{  }r@{}}
    \toprule
      & & \multicolumn{2}{c}{Precision} & \multicolumn{2}{c}{Recall} & \multicolumn{2}{c}{F1} &  \multicolumn{2}{c}{Total Hyps.}  \\
      &  Seeds  &   Procrustes &  SoftSGM  &  Procrustes & SoftSGM  &  Procrustes &  SoftSGM &  Proc. &  SoftSGM\\
    \midrule
    \multirow{6}{*}{\textbf{En-De}} 
& 100  & 2.2  & \textbf{30.2} & 11.2 & \textbf{53.5} & 3.7  & \textbf{38.6} & 24015 & 8516 \\
& 200  & 6.8  & \textbf{34.8} & 33.9 & \textbf{53.3} & 11.3 & \textbf{42.1} & 23515 & 7203 \\
& 500  & 13.9 & \textbf{43.1} & \textbf{69.6} & 55.7 & 23.2 & \textbf{48.6} & 22015 & 5694 \\
& 1000 & 15.9 & \textbf{48.2} & \textbf{79.6} & 57.1 & 26.5 & \textbf{52.3} & 19515 & 4625 \\
& 2000 & 16.8 & \textbf{58.3} & \textbf{83.8} & 62.6 & 28.0 & \textbf{60.4} & 14515 & 3117 \\
& 4000 & 17.2 & \textbf{74.2} & \textbf{86.2} & 74.2 & 28.7 & \textbf{74.2} & 4515 & 903 \\
    \midrule
    \midrule
\multirow{6}{*}{\textbf{Ru-En}} 
& 100  & 2.5  & \textbf{33.3} & 12.6 & \textbf{59.8} &  4.2 & \textbf{42.8} & 19920  & 7150 \\
& 200  & 7.6  & \textbf{38.2} & 38.0 & \textbf{59.4} & 12.7 & \textbf{46.5} & 19420  & 6046 \\
& 500  & 14.1 & \textbf{45.8} & \textbf{70.3} & 59.6 & 23.5 & \textbf{51.8} & 17920  & 4666 \\
& 1000 & 16.0 & \textbf{53.8} & \textbf{80.0} & 59.8 & 26.7 & \textbf{56.6} & 15420  & 3430 \\
& 2000 & 16.8 & \textbf{67.1} & \textbf{83.9} & 67.1 & 28.0 & \textbf{67.1} & 10420  & 2084 \\
& 4000 & 17.1 & \textbf{89.3} & 85.7 & \textbf{89.3} & 28.5 & \textbf{89.3} & 420  &   84 \\ 
    \bottomrule
  \end{tabular}
  \caption{P@5, Recall@5, and F1@5 of Procrustes vs. SGM. ``Total Hyps." = total number of hypotheses.}
  \label{tab:p5-single-results}
\end{table*}

\section{Iterative Experiments}
\label{iter-exps}
\subsection{Methods}
\begin{figure}[htb]
  \centering
  \includegraphics[width=1\linewidth]{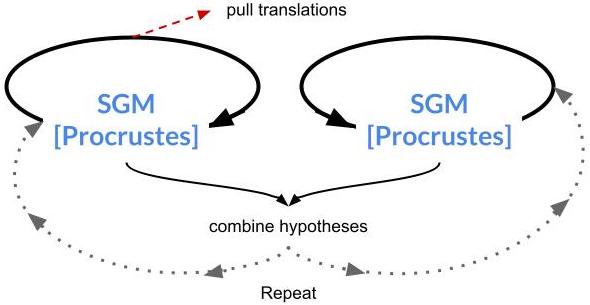}
  \caption{IterSGM [or IterProc]. Run SGM [or Procrustes] in forward \& reverse directions. Combine hypotheses and pass as seeds to SGM [Procrustes]. Pull final translations on last iteration from forward run.}
  \label{fig:iter-algs-img}
\end{figure}

It is popular to use the Procrustes solution iteratively.  One applies the transformation calculated via Procrustes, extracts a dictionary of translation candidates, then uses those as seeds for the next round of Procrustes. We develop an analogous iterative algorithm for SGM.  Figure \ref{fig:iter-algs-img} illustrates the two related approaches. 

We run SGM or Procrustes and extract potential translation pairs in source$\rightarrow$target and target$\rightarrow$source language directions, resulting in two sets of translation hypotheses (one hypothesis per source word from each translation direction). For Procrustes, this extraction is done with CSLS.  We intersect the hypotheses from the two directions, and feed the resulting set back to Procrustes [or SGM] as seeds. We abbreviate the iterative procedures as IterProc and IterSGM. 

The general procedure is: 
\vspace{5pt}
\begin{enumerate}[nolistsep]
\item Run Procrustes [or SGM], forward direction.
\item Run Procrustes [or SGM], reverse direction.
\item Intersect the hypotheses from both directions.
\item Feed the hypotheses into step 1. Repeat.
\end{enumerate}
\vspace{5pt}
Before step 1 for IterSGM, we form the graphs as described in Section \ref{noniter-exps}. For IterProc, we combine the hypotheses in step 3 with the gold seeds, which is unncessary for SGM because seeds are always returned in the hypotheses. 

\begin{table*}[ht]
  \centering
  \begin{tabular}{@{}ll|cc|cc|cc@{}}
    \toprule
     &  & \multicolumn{2}{c}{Add-All} & \multicolumn{2}{c}{Stochastic-Add} & \multicolumn{2}{c}{Active-Learning} \\
        & Seeds & IterProc & IterSGM & IterProc & IterSGM & IterProc & IterSGM\\
    \midrule
    \multirow{6}{*}{En-De}
     & 100 &  \textbf{61.3} & 47.2 & \textbf{62.1} \textcolor{darkgreen}{\textit{ (+0.8)}} &           50.2 \textcolor{darkgreen}{\textit{(+3.0)}} 
                & \textbf{66.1} \textcolor{darkgreen}{\textit{(+4.8)}} & 56.6         \textcolor{darkgreen}{\textit{(+9.4)}}\\
     & 200 &  \textbf{61.5} & 48.2 & \textbf{62.0} \textcolor{darkgreen}{\textit{ (+0.5)}} &           50.8 \textcolor{darkgreen}{\textit{(+2.6)}} 
                & \textbf{66.3} \textcolor{darkgreen}{\textit{(+4.8)}} & 56.7         \textcolor{darkgreen}{\textit{(+8.5)}} \\
     & 500 &  \textbf{62.6} & 52.1 & \textbf{62.8} \textcolor{darkgreen}{\textit{ (+0.2)}} &           52.9 \textcolor{darkgreen}{\textit{(+0.8)}} 
                & \textbf{66.6} \textcolor{darkgreen}{\textit{(+4.0)}} & 58.3         \textcolor{darkgreen}{\textit{(+6.2)}}\\
     & 1000 & \textbf{63.0} & 54.7 & \textbf{63.5} \textcolor{darkgreen}{\textit{ (+0.5)}} &           54.8 \textcolor{darkgreen}{\textit{(+0.1)}} 
                & \textbf{67.3} \textcolor{darkgreen}{\textit{(+4.3)}} & 59.5         \textcolor{darkgreen}{\textit{(+4.8)}}\\
     & 2000 & \textbf{65.2} & 61.4 & \textbf{65.2} \textcolor{gray}{\textit{ (+0.0)}} &           61.7 \textcolor{darkgreen}{\textit{(+0.3)}} 
                & \textbf{69.1} \textcolor{darkgreen}{\textit{(+3.9)}} & 65.6         \textcolor{darkgreen}{\textit{(+4.2)}}\\
     & 4000 & 71.3 & \textbf{74.2} &          71.7 \textcolor{darkgreen}{\textit{ (+0.4)}} &  \textbf{74.4} \textcolor{darkgreen}{\textit{(+0.2)}} 
                & 74.6          \textcolor{darkgreen}{\textit{(+3.3)}} & \textbf{75.4} \textcolor{darkgreen}{\textit{(+1.2)}} \\
    \midrule  
    \midrule  
     \multirow{6}{*}{Ru-En}
     & 100 &  \textbf{62.4} & 51.6 & \textbf{62.7} \textcolor{darkgreen}{\textit{(+0.3)}} &  56.3   \textcolor{darkgreen}{\textit{(+4.7)}} 
                & \textbf{71.0} \textcolor{darkgreen}{\textit{(+8.6)}}  & 62.5          \textcolor{darkgreen}{\textit{(+10.9)}}    \\
     & 200 &  \textbf{62.4} & 53.7 & \textbf{63.1} \textcolor{darkgreen}{\textit{(+0.7)}} &  56.4   \textcolor{darkgreen}{\textit{(+2.7)}} 
                & \textbf{71.1} \textcolor{darkgreen}{\textit{(+8.7)}}  & 61.9          \textcolor{darkgreen}{\textit{(+8.2)}}  \\
     & 500 &  \textbf{63.7} & 56.1 & \textbf{63.7} \textcolor{gray}{\textit{(+0.0)}}                        &  58.0   \textcolor{darkgreen}{\textit{(+1.9)}} 
                & \textbf{71.3} \textcolor{darkgreen}{\textit{(+7.6)}}  & 63.1          \textcolor{darkgreen}{\textit{(+7.0)}} \\
     & 1000 & \textbf{64.0} & 58.1 & \textbf{64.0} \textcolor{gray}{\textit{(+0.0)}}                        &  60.3   \textcolor{darkgreen}{\textit{(+2.2)}} 
                & \textbf{71.2} \textcolor{darkgreen}{\textit{(+7.2)}}  & 66.4          \textcolor{darkgreen}{\textit{(+8.3)}} \\
     & 2000 & 66.1 & \textbf{67.1} &          65.7 \textcolor{red}{\textit{( -0.4)}} & \textbf{68.2} \textcolor{darkgreen}{\textit{(+1.1)}} 
                & \textbf{72.3} \textcolor{darkgreen}{\textit{(+6.2)}}  & 71.0          \textcolor{darkgreen}{\textit{(+3.9)}} \\
     & 4000 & 69.0 & \textbf{89.3} &          69.0 \textcolor{gray}{\textit{(+0.0)}}                  & \textbf{89.3} \textcolor{gray}{\textit{(+0.0)}}                        
                & 72.6          \textcolor{darkgreen}{\textit{(+3.6)}}  & \textbf{89.3} \textcolor{gray}{\textit{(+0.0)}} \\ 
    \bottomrule
  \end{tabular}
  \caption{P@1 of IterProc vs. IterSGM. Add-All runs for 10 iterations, seeding subsequent iterations with the intersection of hypotheses from forward and reverse directions. For Stochastic-Add, seeds are fed in up to 100 at a time until all are used. In parentheses is the improvement over Add-All.}
  \label{iter-results}
\end{table*}
How one select seeds in Step 4 for subsequent rounds is important.  We try three variations:
\paragraph{Add-All} Itersect hypotheses from forward and reverse directions. All become seeds for the next round, for $N$ total rounds. Advantage: all correct pairs are passed to the next iteration. Disadvantage: all incorrect hypotheses are, too.
\paragraph{Stochastic-Add} Add up to $H$ new hypotheses each iteration; For iteration two, $H$ random hypotheses from the intersection are chosen and added to the gold seeds for the forward direction. A separate random selection is taken for the reverse direction. The next round, 2$H$ random hypotheses are chosen. This continues until all hypotheses are used.\footnote{If not enough seeds in the intersection, all are used.} This setting was designed to encourage robustness and improve accuracy by minimizing the number of erroneous seeds passed to subsequent rounds, to allow for recovery from mistakes.
As distinct subsets are passed to forward and reverse directions, we encourage solutions of the runs to also be different, increasing output diversity. When intersecting the hypotheses, we aim to select pairs which are most likely to be correct---having two different solutions agree increases confidence that the induced pairs are correct, and selecting only a small subset allows recovery from mistakes. This is particularly important for SGM, where incorrect seeds are repeated in the output. In passing a subset to the next round, some incorrect pairs are dropped, and the model gets another chance to induce translations with a (presumably) stronger model. The stochasticity builds in robustness. 
\paragraph{Active-Learning} Seeds may also be added in an active learning fashion (``human-in-the-loop"). To simulate a human judging hypothesis quality, we use the union of hypotheses from forward and reverse directions and pass only correct hypotheses as seeds for the next iteration. 

\vspace{5pt}
For Add-All and active learning experiments, we run for ten iterations ($N=10$). For Stochastic-Add, $H=100$. Tuning $H$ is for future work. 

\subsection{Results} 
Results for IterProc and IterSGM are in Table \ref{iter-results}. In parentheses is the raw improvement over Add-All. Unlike the single runs of Procrustes and SGM from Table \ref{tab:single-results}, IterProc outperforms IterSGM in all scenarios with 1000 or less seeds, and for En-De with 2000 seeds. Stochastic-Add outperforms Add-All in nearly all experiments. Because SGM is more sensitive to input seeds than Procrustes, it particularly benefits from the stochastic setup which minimizes its exposure to incorrect input seeds and allows recovery from mistakes. Both IterProc and IterSGM benefit from active learning, showing the improved performance that may be achieved from human-in-the-loop.
\begin{figure*}[t]
  \centering
  \includegraphics[width=1\linewidth]{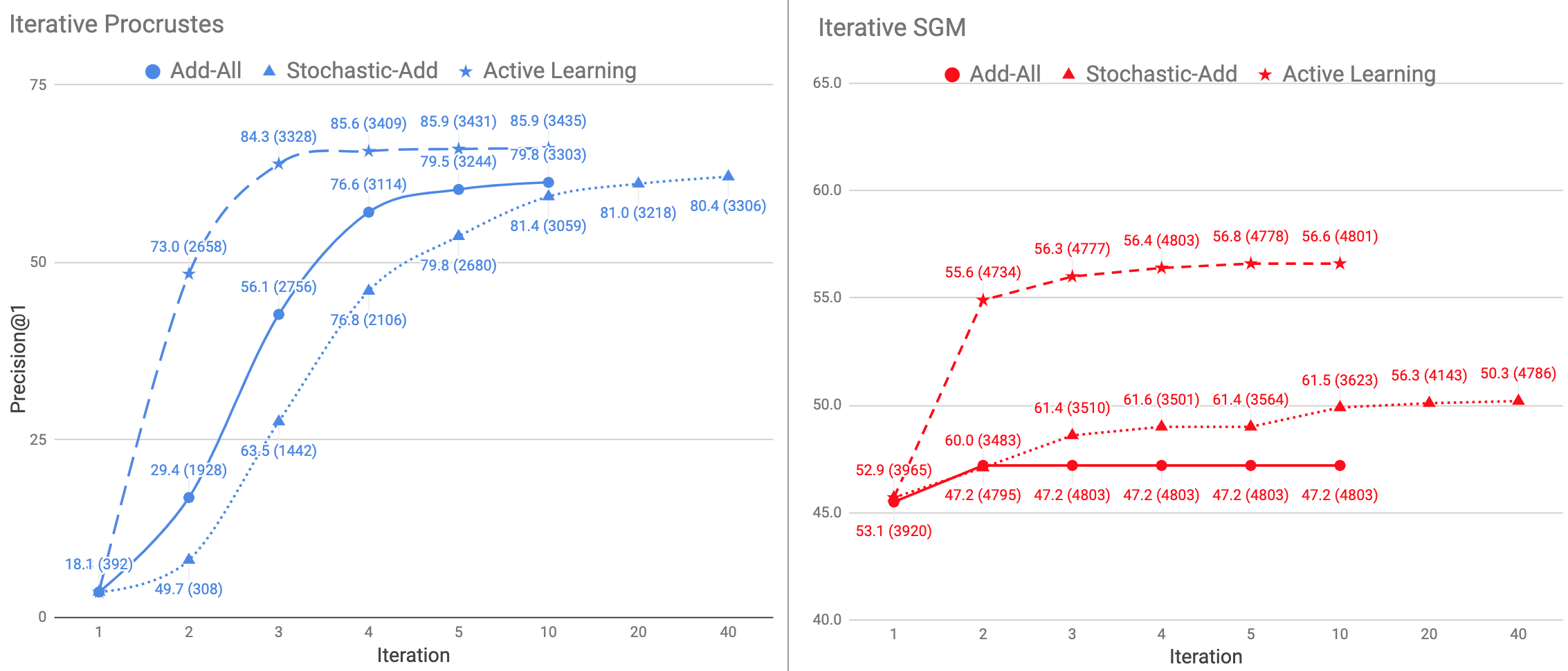}
  \caption{P@1 of iterative methods, by iteration (En-De, 100 seeds). Y-axis has p@1 for the forward run of Procrustes [SGM]. Seeds for subsequent iterations are drawn from the intersection of forward and reverse runs. Size and precision of this intersection is labeled above each point as ``Precision (Num Hyps)". Add-All adds all hypotheses in the intersection as seeds to the next iteration. Stochastic-Add adds random samples of up to 100 new hypotheses per iteration. Active Learning adds all \textbf{correct} hypotheses.}
  \label{fig:bothiter-graphs}
\end{figure*}

Figure \ref{fig:bothiter-graphs} has p@1 for IterProc vs. IterSGM during training (En-De, 100 seeds). Each data point has the number of hypotheses in the intersection of forward and reverse runs, and the precision of the intersection [Precision (Hyps.)]. IterProc dramatically underperforms SGM initially but quickly recovers. IterSGM stays roughly consistent throughout iterations. Precision of IterProc rapidly improves, but stays roughly the same for IterSGM. The number of hypotheses in the intersection is smaller for IterProc, suggesting that forward and reverse directions disagree more, but the hypotheses that they do agree upon are more precise. 

\vspace{5pt}
The results in this section and the previous suggest that Procrustes and SGM have complementary strengths. While a single run of Procrustes struggles to align word embedding spaces with little supervision, it recovers when run iteratively. Conversely, one run of SGM dramatically outperforms one run of Procrustes with low number of seeds but does not improve much with iterations. 

\section{System Combination Experiments}
\label{combo-exps}
\subsection{Method}
\begin{figure}[htb]
  \centering
  \includegraphics[width=1\linewidth]{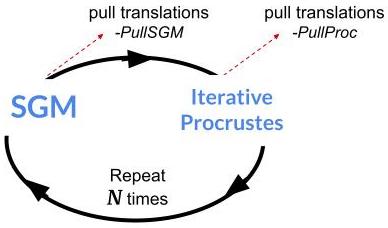}
  \caption{Combined cyclic system. SGM and Add-All IterProc are interspersed. Begin and end anywhere on the cycle. (1) Run SGM [or IterProc] in forward/reverse direction. (2) Intersect hypotheses and pass to forward/reverse IterProc [SGM] as seeds. (3) Pull final translations after $N$th cycle from forward SGM [IterProc].}
  \label{fig:combined}
\end{figure}

We create a combined system to see whether both methods together can outperform either alone, shown in Figure \ref{fig:combined}. For simplicity of implementation, we use Add-All IterProc and single runs of SGM. Here, Procrustes and SGM feed off one another to iteratively improve the solution. The combined system is cyclic---one may choose where to begin and end, with differing effect. There are two main components and a hypothesis extraction step:
\begin{enumerate}[nolistsep,label={\Alph*.}]
\item \textbf{SGM} Run in forward and reverse directions. Intersect hypotheses and pass to next step.
\item \textbf{IterProc} Run for $I_\text{proc}$ iterations. 
\item \textbf{Hypothesis Extraction} Pull translation pairs from a forward run of SGM (\mbox{-PullSGM}) or IterProc (\mbox{-PullProc}). This results in one hypothesis for each source word. 
\end{enumerate}
We set $I_\text{proc} = 5$ and $N=10$. We start from IterProc and pull results either from IterProc (Start: IterProc \mbox{-PullProc}) or SGM (Start: IterProc \mbox{-PullSGM}) on the final loop of the cycle. We repeat the experiments starting from SGM (Start: SGM). 

\subsection{Results}
Results for the combined cyclic system are in Table \ref{tab:combined-results}. The ``Previous Best" column has the best performance from previous experiments (excluding active learning).  For all seed levels, the cyclic system can equal or outperform the previous best performance from earlier experiments with single and iterative Procrustes or SGM.   

\begin{figure}[t]
  \centering
  \includegraphics[width=1\linewidth]{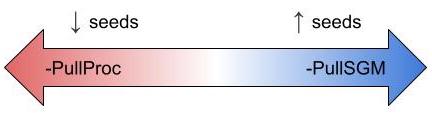}
  \caption{Whether pulling from IterProc or SGM is preferred in the combined cyclic system depends on the number of seeds, with IterProc preferred with a low number of seeds, and SGM preferred with more.}
  \label{fig:continuum}
\end{figure}

\begin{table*}[ht]
\centering
\begin{tabular}{l|l|c||cc|cc}
  \toprule
& & & \multicolumn{4}{c}{Combination Methods} \\
& & & \multicolumn{2}{c}{\underline{\mbox{-PullProc}}} & \multicolumn{2}{c}{\underline{\mbox{-PullSGM}}} \\
& Seeds & Prev. Best & \textit{Start: IterProc} & \textit{Start: SGM} &  \textit{Start: IterProc} & \textit{Start: SGM} \\
\hline
\multirow{6}{*}{En-De} 
& 100  & 62.1           & \textbf{62.2} & 62.1 & 59.7          & 59.5          \\
& 200  & 62.0           & \textbf{62.8} & 62.6 & 60.4          & 60.4          \\
& 500  & 62.8           & 63.5 & \textbf{63.8} & 62.1          & 62.0          \\
& 1000 & 63.5           & 63.9 & \textbf{64.2} & 63.0          & 63.7          \\
& 2000 & 65.2           & 66.7          & 66.7          & \textbf{69.7} & 69.0 \\
& 4000 & 74.4          & 73.2          & 73.2          & \textbf{79.7} & 79.2 \\
    \midrule  
    \midrule  
\multirow{6}{*}{Ru-En}   
& 100  & 62.7           & 63.9 & \textbf{64.0} & 61.7          & 62.0          \\
& 200  & 63.1           & \textbf{64.5} & 64.3 & 62.6          & 63.1          \\
& 500  & 63.7           & \textbf{65.3} & \textbf{65.3} & 64.0          & 64.3          \\
& 1000 & 64.0           & \textbf{66.8} & 66.4 & \textbf{66.8} & 66.4 \\
& 2000 & 68.2          & 69.4          & 69.5          & 72.9 & \textbf{73.1} \\
& 4000 & \textbf{89.3} & 77.4          & 77.4          & \textbf{89.3} & \textbf{89.3} \\
    \bottomrule
  \end{tabular}
  \caption{P@1 for combined cyclic method (Figure \ref{fig:combined}. One may begin from either IterProc (``Start: Procrustes") or SGM (``Start: SGM"), and may pull final hypotheses from either Procrustes (``\mbox{-PullProc}") or SGM (``\mbox{-PullSGM}"). ``Prev. Best" is best result from previous experiments (excluding active learning). Bold is best overall.}
  \label{tab:combined-results}
\end{table*}
Looking down the ``\mbox{-PullProc}" columns, we discover that it hardly matters whether we begin the cycle with IterProc or SGM. The same is true for ``\mbox{-PullSGM}". Whether \mbox{-PullProc} or -PullSGM is preferred appears to be on a continuum, depicted in Figure \ref{fig:continuum}. For a low seed count, \mbox{-PullProc} is preferred, with the effect more pronounced as seed size diminishes. Conversely, \mbox{-PullSGM} is increasingly preferred as seed set size increases.

\section{Discussion \& Future Directions}
Though much work in BLI takes a Euclidean view and elicits solutions via methods such as solutions to the generalized Procrustes problem, BLI may also be viewed as a graph-based problem. Parts of the graph-based view have appeared in existing work, but no one has yet to compare the different framings in the context of BLI. We perform this analysis for two high-resource language pairs with well-trained embeddings from Wikipedia, in a restricted data context. Under our experimental settings, we find that:
\begin{enumerate}[nolistsep]
\item Procrustes-based methods and SGM behave differently under differing contexts (namely, the amount of available seeds), so either may be favorable given the specific data context. SGM appears favorable with less seeds.
\item SGM can be run iteratively, but does not improve as rapidly as Iterative Procrustes. Both benefit from stochasticity, and active learning can provide strong improvement.
\item Procrustes and SGM can be effectively combined to outperform either alone.
\end{enumerate}

Our work has limitations which should be addressed by future analyses. We use clean, well-trained embeddings from the same domain. Previous work has shown Procrustes to struggle with poorly-trained and low-resource word embedding spaces, and for well-trained embeddings in mismatched domains \cite[e.g.,][]{marchisio-etal-2020-unsupervised}. In these cases, SGM might benefit from a different distance metric. A detailed analysis should be performed when data is many-to-many, as translation is naturally a many-to-many task. One might revisit word vectors based on co-occurrence statistics. The size of training and test sets should be increased, as the presence of more synonyms/antonyms and other ``distractor" words may elicit different behavior. There are computational considerations as we scale-up, particularly for SGM. 

\section{Related Work}
Matching words using vector representations began with vectors based on co-occurrence statistics. \citet{rapp1995} and \citet{fung-1995-compiling} induce bilingual lexica based on the principle that words that frequently co-occur in one language have translations that co-occur frequently in another. \citet{Diab-Finch-2000} extend this by measuring similarity between words based on co-occurrence vectors and matching words across language by preserving these similarities. 
\citet{mikolov2013} are the first to perform BLI over word embeddings, estimating the transformation matrix using stochastic gradient descent. Most recent work solves a variation of the generalized Procrustes problem \cite[e.g.,][]{conneau-lample-2018, artetxe-etal-2016-learning, artetxe-etal-2017-learning, patra-etal-2019-bilingual, artetxe-etal-2018-robust, doval2018improving, joulin2018loss,jawanpuria2019learning, alvarez-melis-jaakkola-2018-gromov}. \citet{zhang-etal-2020-overfitting} learn a mapping that overfits to training pairs thus enforcing ``hard-seeding", while \citet{ruder-etal-2018-discriminative} enforce a one-to-one constraint on the output for BLI.

Some BLI work uses graph based methods implicitly or explicitly. \citet{artetxe2018robust} form an initial solution with similarity matrices and refine with iterative Procrustes. \citet{grave-etal-2019-training} optimize ``Procrustes in Wasserstein Distance", employing a quadratic assignment formulation and the Frank-Wolfe method. \citet{ren-etal-2020-graph} form CSLS similarity matrices, iteratively extract cliques, and map with Procrustes. \citet{gutierrez2017low} create a weighted graph of translation candidates then create word vectors with Node2Vec \cite{grover2016node2vec}. \citet{wushouer2013inducing} use graphs for a source, target, and pivot language to iteratively extract translation pairs based on heuristics. Our active learning approach is inspired by \citet{yuan-etal-2020-interactive-refinement}.

\section{Conclusion}
We perform the first detailed analysis of the consequences of framing BLI either as a Euclidean problem solved by the common Procrustes solution with nearest-neighbor search, or as a graph-based matching problem solved with SGM. We show that each performs differently under different data contexts, with SGM preferred with low amounts of seeds. We compare iterative versions of SGM and Procrustes, and find that stochasticity benefits both. Finally, we create a combined system that outperforms individual Procrustes and SGM approaches. 

\section*{Acknowledgements}
We thank our anonymous reviewers for their comments. This material is based upon work supported by the United States Air Force under Contract No. FA8750‐19‐C‐0098.  Any opinions, findings, and conclusions or recommendations expressed in this material are those of the author(s) and do not necessarily reflect the views of the United States Air Force and DARPA. This work was supported in part by the US Defense Advanced Research Projects Agency under the D3M program administered through contract FA8750-17-2-0112.

\bibliographystyle{acl_natbib}
\bibliography{anthology, emnlp2021}

\clearpage

\appendix

\section{Appendix for: An Analysis of Euclidean vs. Graph-Based Framing for Bilingual Lexicon Induction from Word Embedding Spaces}

\subsection{Mathematical Notation}
$\oplus$ is the direct sum of matrices:
$$I_s \oplus P = \begin{bmatrix}
I_s & 0 \\
0 & P
\end{bmatrix}$$

\subsection{Simple Example of Seeded Graph Matching}

Recall the constrained optimization objective for seeded graph matching:
\begin{equation*}
\underset{P \in \Pi_{n-s}}{\arg\min}\lVert G_x - (I_s \oplus P)G_y(I_s \oplus P)^T \rVert^2_F
\end{equation*}
Let $x_1, x_2, x_3, x_4 \in X$ and $y_1, y_2, y_3, y_4 \in Y$. To more clearly see the effect, each of the vectors is orthogonal to all others but not unit length. We create the graphs as below, where each $G_{x_{ij}} = \langle x_i, x_j\rangle$ (equivalently for $G_{y_{ij}} \in G_y$). We take $(x_1, y_1)$ as a seed.
$$ G_x =
\begin{bmatrix}
2 & 0 & 0 & 0 \\
0 & 2 & 0 & 0 \\
0 & 0 & 3 & 0 \\
0 & 0 & 0 & 4
\end{bmatrix}, 
G_y =
\begin{bmatrix}
1 & 0 & 0 & 0 \\
0 & 3 & 0 & 0 \\
0 & 0 & 4 & 0 \\
0 & 0 & 0 & 2
\end{bmatrix}
$$
To minimize $G_x - G_y$, we swap $y_2$ to $y_3$, $y_3$ to $y_4$, and $y_4$ to $y_2$ using P as below:
$$
P =
\begin{bmatrix}
0 & 1 & 0 \\
0 & 0 & 1 \\
1 & 0 & 0
\end{bmatrix}
$$
Let ${G'}_y = (I_s \oplus P)G_y(I_s \oplus P)^T$:
\begin{gather*}
{G'}_y = 
\begin{bmatrix}
1 & 0 & 0 & 0 \\
0 & 0 & 1 & 0 \\
0 & 0 & 0 & 1 \\
0 & 1 & 0 & 0
\end{bmatrix}\begin{bmatrix}
1 & 0 & 0 & 0 \\
0 & 3 & 0 & 0 \\
0 & 0 & 4 & 0 \\
0 & 0 & 0 & 2
\end{bmatrix}\begin{bmatrix}
1 & 0 & 0 & 0 \\
0 & 0 & 0 & 1 \\
0 & 1 & 0 & 0 \\
0 & 0 & 1 & 0
\end{bmatrix}
\\
= \begin{bmatrix}
1 & 0 & 0 & 0 \\
0 & 2 & 0 & 0 \\
0 & 0 & 3 & 0 \\
0 & 0 & 0 & 4
\end{bmatrix}
\end{gather*}
We note that this choice for ${G'}_y$ (and therefore $P$) minimizes Equation \ref{1}. The solutions we extract as translation pairs from $G_x$ and $G_y$ are therefore $(x_1, y_1), (x_2, y_4), (x_3, y_2), (x_4, y_3)$.

\subsection{Seeded Graph Matching}
\label{sgm}
This section describes Seeded Graph Matching \cite{fishkind2019seeded}.
Let $G_x, G_y \in \mathbb{R}^{n\times n} $ be graphs representing the relationships between words in word embedding spaces $X, Y \in \mathbb{R}^{n\times d}$, respectively. We use cosine similarity as the measure of distance when weighting the edges, and therefore the resulting graphs are undirected and symmetric. To form $G_x$, we may normalize the embeddings in $X$ so that $G_x = XX^T$.  We create $G_y$ similarly. 

Assume seeds $\{(x_1, y_1), (x_2, y_2),... (x_s, y_s)\}$ are given. We formulate this constrained optimization problem as below:
\begin{equation}
\label{1}
\underset{P \in \Pi_{n-s}}{\arg\min}\lVert G_x - (I_s \oplus P)G_y(I_s \oplus P)^T \rVert^2_F
\end{equation}
We understand $(I_s \oplus P)G_y(I_s \oplus P)^T$ as the attempt to ``move" the rows/columns of the graph $G_y$ such that its rows/columns are in the same order as $G_x$, which is equivalent to relabeling the edges in $G_y$. Rows/columns in $G_x$ and $G_y$ after reordering that have the same index are then extracted as translations of one another. 

We rearrange Equation \ref{1} to be more tractable for optimization. Letting ${G'}_y = (I_s \oplus P)G_y(I_s \oplus P)^T$, we perform the below:\footnote{Recall the definition of norm: $||x|| = \sqrt{\langle x, x \rangle}$, properties of the inner product, the trace definition of Frobenius inner product whereby $\langle A, B \rangle_F = tr(B^TA)$, and the fact that a permutation matrix's transpose is its inverse.}
\begin{gather*}
\underset{P \in \Pi_{n-s}}{\arg\min}\langle G_x - {G'}_y, G_x - {G'}_y \rangle_F \\
\quad = \underset{P \in \Pi_{n-s}}{\arg\min} ||G_x||^2_F + ||G_y||^2_F - 2 \cdot tr(G_x^T {G'}_y) \\
\qquad = \underset{P \in \Pi_{n-s}}{\arg\min} \;  -2 \cdot tr(G_x^T {G'}_y) \\
\qquad = \underset{P \in \Pi_{n-s}}{\arg\max} \;\; tr(G_x^T (I_s \oplus P)G_y(I_s \oplus P)^T) \\
\end{gather*}
Because the original objective is non-convex, the constraint on P is relaxed to being a doubly-stochastic matrix\footnote{All rows/columns sum to 1.} $P \in D_{n-s}$, which is the convex hull of the set of permutation matrices (Birkhoff-Von Neumann Theorem). The resulting optimization objective is thus:
\begin{equation}
\qquad = \underset{P \in D_{n-s}}{\arg\max} \;\; tr(G_x^T (I_s \oplus P)G_y(I_s \oplus P)^T) 
\end{equation}

\end{document}